\documentclass[conference]{IEEEtran}
\usepackage{amssymb}
\usepackage{cite}
\usepackage{graphicx}
\usepackage{epstopdf}
\graphicspath{{./figures/}}

\usepackage[cmex10]{amsmath}
\usepackage{algorithmic}
\usepackage{array}
\ifCLASSOPTIONcompsoc
 \usepackage[caption=false,font=normalsize,labelfont=sf,textfont=sf]{subfig}
\else
 \usepackage[caption=false,font=footnotesize]{subfig}
\fi
\usepackage{dsfont}
\usepackage{enumerate}
\usepackage{enumitem}
\usepackage{bbm}
\usepackage{gensymb}

\setlist[description]{leftmargin=0cm,labelindent=0cm}

\def\bx{{\mathbf x}}
\def\bX{{\mathbf X}}

\newtheorem{theorem}{Theorem}[section]

\newcommand{\qed}{\nobreak \ifvmode \relax \else
      \ifdim\lastskip<1.5em \hskip-\lastskip
      \hskip1.5em plus0em minus0.5em \fi \nobreak
      \vrule height0.75em width0.5em depth0.25em\fi}
\def\Reals{{\mathbb R}}
\def\btheta{{\boldsymbol \theta}}
\def\bmu{{\boldsymbol \mu}}
\def\bomega{{\boldsymbol \omega}}
\def\bgamma{{\boldsymbol \gamma}}
\def\bQ{{\mathbf{Q}}}

\def\argmax{{\mathbf{amax}}}

\begin{document}

\title{Parameter estimation in spherical symmetry groups}

\author{
\IEEEauthorblockN{Yu-Hui Chen\IEEEauthorrefmark{1},
Dennis Wei\IEEEauthorrefmark{2},
Gregory Newstadt\IEEEauthorrefmark{3}, Marc DeGraef\IEEEauthorrefmark{4}, 
Jeffrey Simmons\IEEEauthorrefmark{5} and
Alfred Hero\IEEEauthorrefmark{1}}
\IEEEauthorblockA{\IEEEauthorrefmark{1}University of Michigan, Ann Arbor, MI USA}
\IEEEauthorblockA{\IEEEauthorrefmark{2}IBM Watson Research Center, Yorktown Heights, NY USA}
\IEEEauthorblockA{\IEEEauthorrefmark{3}Google Inc., Pittsburgh, PA USA}
\IEEEauthorblockA{\IEEEauthorrefmark{4}Carnegie Mellon University Pittsburgh, PA USA}
\IEEEauthorblockA{\IEEEauthorrefmark{5}US Air Force Research Laboratory (AFRL), Dayton, OH USA}
}

\maketitle

\begin{abstract}
This paper considers statistical estimation problems where the probability distribution of the observed random variable is invariant with respect to actions of a finite topological group. It is shown that any such distribution must satisfy a restricted finite mixture representation. When specialized to the case of distributions over the sphere that are invariant to the actions of a finite spherical symmetry group $\mathcal G$, a group-invariant  extension of the Von Mises Fisher (VMF) distribution is obtained. The $\mathcal G$-invariant VMF is parameterized by location and scale parameters that specify the distribution's mean orientation and its concentration about the mean, respectively. Using the restricted finite mixture representation these parameters can be estimated using an Expectation Maximization (EM) maximum likelihood (ML) estimation algorithm. This is illustrated for the problem of mean crystal orientation estimation under the spherically symmetric group associated with the crystal form, e.g., cubic or octahedral or hexahedral. Simulations and experiments  establish the advantages of the extended VMF EM-ML estimator for data acquired by Electron Backscatter Diffraction (EBSD) microscopy of a polycrystalline Nickel alloy sample. \footnote{AOH would like to acknowledge financial support from USAF/AFMC grant FA8650-9-D-5037/04 and AFOSR grant FA9550-13-1-0043. MDG would like to acknowledge financial support from AFOSR MURI grant FA9550-12-1-0458.}
\end{abstract}

\ifCLASSOPTIONpeerreview
\begin{center} \bfseries EDICS Category: 3-BBND \end{center}
\fi
%
\IEEEpeerreviewmaketitle

\section{Introduction}
\label{sec:intro}
This paper considers estimation of parameters of distributions whose domain is a particular non-Euclidean geometry: a topological space divided into $M$ equivalence classes by actions of a finite spherical symmetry group. A well known example of a finite spherical symmetry group is the point group in 3 dimensions describing the soccer ball, or football, with truncated icosahedral symmetry that also corresponds to the symmetry of the Carbon-60 molecule.  This paper formulates a general approach to parameter estimation in distributions defined over such domains. First we establish a restricted finite mixture representation for probability distributions that are invariant to actions of any topological group. This representation has the property that the number of mixture components is equal to the order of the group, the distributions in the mixture are all parameterized by the same parameters,  and the mixture coefficients are all equal. This is practically significant since many reliable algorithms have been developed for parameter estimation when samples come from finite mixture distributions.

We illustrate the power of the representation for an important problem in materials science: analysis  of mean orientation in polycrystals.  Crystal orientation characterizes properties of materials including electrical conductivity and thermal conductivity. Mechanical properties, such as, stiffness, elasticity, and deformability, can also depend on the distribution of crystal orientations over the material. Thus accurate estimation of crystal orientation is useful for materials evaluation, testing and prediction.

The mean orientation of the crystal, characterized by its Euler angles, can only be specified modulo a set of angular rotations determined by the symmetry group associated with the specific type of crystal. This multiplicity of equivalent Euler angles complicates the development of reliable mean orientation estimators. By extending the Von Mises Fisher (VMF) model under the proposed finite mixture representation,  and applying the expectation maximization (EM) maximum likelihood (ML) algorithm for mixtures, we obtain an accurate iterative estimator of the mean Euler angle parameter and angular concentration parameter of the extended VMF distribution. Specifically, the VMF extension is accomplished as follows. We start with the standard VMF model, which is a density parameterized by location (angle mean) and scale (angle concentration) defined over the $p$-dimensional sphere \cite{mardia_directional_1999}. In this model, a point on the sphere is specified by its direction vector, and the angle between two vectors is the arc-cosine of the normalized inner product between them. The spherical symmetry group extension is accomplished by applying the mixture representation to the standard VMF distribution using the group of quaternion rotation matrices.

The performance of the proposed EM-ML orientation estimator is evaluated by simulation and compared to two other angle estimators. The ML orientation estimator is then illustrated on EBSD data collected from a Nickel alloy whose crystal form induces the $m\overline{3}m$ cubic point symmetry group. We establish that the ML orientation estimator results in significantly improved estimates of the mean direction in addition to providing an accurate estimate of concentration about the mean.

The paper is organized as follows. Section~\ref{sec:group-invariant} describes group invariant random variables and gives the mixture representation for their densities. Section \ref{sec:spherical_symmetry_group} specializes to random variables invariant relative to actions of the spherical symmetry group and develops the $\mathcal G$-invariant VMF distribution along with  EM-ML parameter estimator. The crystallography application is presented in section~\ref{sec:app_crystal_orientation_estimation} along with experimental comparisons. Section~\ref{sec:conclusion} has concluding remarks.

\section{Group-invariant random variables}
\label{sec:group-invariant}
\def\bfx{{\mathbf x}}
Consider a finite topological group $\mathcal G=\{G_1, \ldots, G_M\}$ of $M$ distinct actions on a topological space $\mathcal X$,  $G_i: \mathcal X \rightarrow \mathcal X$ and a binary operation "*" defining the action composition $G_i * G_j$, denoted $G_i G_j$. $\mathcal G$ has the properties that composition of multiple actions is associative, for every action there exists an inverse action, and there exists an identity action \cite{birkhoff_brief_1963}.  A real valued function  $f(\bx)$ on  $\mathcal X$ is said to be invariant under $\mathcal G$ if: $f(G\bx)=f(\bx)$ for $G\in \mathcal G$.   Let $\bX$ be a random variable defined on $\mathcal X$.  We have the following theorem for the probability density $f(\bx)$ of $\bX$.
\begin{theorem}
The density function $f: \mathcal X\rightarrow \Reals$ is invariant under $\mathcal G$ if and only if
\begin{eqnarray}
f(\bx)= \frac{1}{M} \sum_{i=1}^M f(G_i\bx).
\label{eq:representation}
\end{eqnarray}
\label{thm:1}
\end{theorem}

\noindent{\em Proof:} If (\ref{eq:representation}) holds then $f(G\bx)=M^{-1} \sum_{i=1}^M f(G_i G \bx)$. Since $\mathcal G$ is a group $\mathcal G G=\mathcal G$ so that  $M^{-1} \sum_{i=1}^M f(G_i G \bx)=M^{-1} \sum_{j=1}^M f(G_j\bx)$ and $f(G\bx)=f(\bx)$. On the other hand, if $f(G\bx)=f(\bx)$ then $\frac{1}{M} \sum_{i=1}^M f(G_i\bx)=\frac{1}{M} \sum_{i=1}^M f(\bx)=f(\bx)$.
\qed

Theorem \ref{thm:1} says that any density $f(\bx)$ that is invariant under group $\mathcal G$ can be represented as a finite mixture of its translates $f(G_i\bx)$ under the group's actions $G_i \in \mathcal G$. This simple result has important implications on $\mathcal G$-invariant density estimation and parameter estimation. In particular it can be used to construct maximum likelihood estimators for parametric densities and kernel density estimators of non-parametric $\mathcal G$-invariant densities with finite sample guaranteed performance.

To illustrate the non-parametric case, assume that  $\mathcal X$ has topological dimension $d$ with Lebesgue $\mathcal G$-invariant density $f(\bx)$. Define the symmetric non-negative second order kernel function $\phi:\mathcal X \rightarrow \Reals$, i.e.,  $\phi(\bx)\geq 0$, $\phi(\bx)=\phi(\|\bx\|,0,\ldots,0)$, $\int \phi(\bx) d\bx =1$, and $\int \|\bx\|^2 \phi(\bx) d\bx<\infty$. For the finite group $\mathcal G$, define the $\mathcal G$-invariant kernel function $K(\bx)=M^{-1} \sum_{i=1}^M \phi(G_i\bx)$. Given a realization $\{\bx_i\}_{i=1}^n$ of $n$ i.i.d. samples from $f$ define the kernel density estimator $\hat{f}_{h}(\bx)= n^{-1}\sum_{i=1}^n K\left(\frac{\bx-\bx_i}{h}\right)$. Assume that $h_n$ is a sequence of kernel widths that satisfies $\lim_{n\rightarrow \infty} h_n =0$ while $\lim_{n\rightarrow \infty} h_n n^d =\infty$. Then, if $f$ is smooth, using Thm~\ref{thm:1} and concentration results from \cite{devroye_combinatorial_2001}, it can be shown that as $n$ goes to infinity
$$E[\|f-\hat{f}_{h_n}\|] = O(n^{-2/(4+d)}), $$
where   $\|f-\hat{f}_{h_n}\|=\int |f(\bx)-\hat{f}_{h_n}(\bx)|d\bx $ is the $\ell 1$ norm difference.

For the parametric case, let $h(\bx;\btheta)$ be a density on $\mathcal X$ that is parameterized by a parameter $\btheta$ in a parameter space $\Theta$. We extend  $h(\bx;\btheta)$ to a $\mathcal G$-invariant density $f$ by using Thm. \ref{thm:1}, obtaining:
\begin{eqnarray}
f(\bx;\btheta)=\frac{1}{M} \sum_{i=1}^M h_i(\bx;\btheta),
\label{eq:SSG}
\end{eqnarray}
where $h_i(\bx;\btheta)=h(G_i\bx;\btheta)$. This density is of the form of a finite mixture of densities $h_i(\bx;\btheta)$ of known parametric form where the mixture coefficients are all identical and equal to $1/M$. Maximum likelihood (ML) estimation of the parameter $\btheta$ from an i.i.d. sample $\{\bX_i\}_{i=1}^n$ from any $\mathcal G$-invariant density $f$ can now be performed using finite mixture model methods \cite{mclachlan_finite_2004} such as the  Expectation-Maximization (EM) algorithm~\cite{dempster_maximum_1977} or the restricted Boltzman machine (RBM) \cite{sohn_efficient_2011}.

\section{ML within a spherical symmetry group}
\label{sec:spherical_symmetry_group}

In this section we specialize to estimation of parameters for the case that the probability density is on a sphere and is invariant to actions in a spherical symmetry group. In Sec.~\ref{sec:app_crystal_orientation_estimation} this will be applied to a crystallography example under a Von-Mises-Fisher likelihood model for the mean crystal orientation. The measured and mean orientations can be represented in three equivalent ways.

\begin{description}
  \item[Euler angles $\mathcal{E}$:] The orientation  is defined by a set of three successive rotations of a reference unit vector about the specified axes~\cite{eberly_euler_2008}. Denote the Euler angles as $\mathbf{e} = (\alpha, \beta, \mathbf{\gamma})\in\mathcal{E}$, where $\alpha, \mathbf{\gamma}\in[0, 2\pi]$ and $\beta\in[0,\pi]$.
  \item[Quaternion $\mathcal{Q}$:]  The quaternion representation describes the orientation as a 4D vector on the 3D sphere \cite{altmann_rotations_2005}: $\mathbf{q}=(q_1,q_2,q_3,q_4)\in\mathcal{Q}$, where $\|\mathbf{q}\|=1$. The main advantage of this representation is that any rotation of $\mathbf q$ is simply accomplished via left multiplication by a $4\times 4$ orthogonal matrix $\bQ$ called a quaternion matrix.
  \item[Rodrigues Vector $\mathcal{D}$:]  The Rodrigues vector describes the orientation by rotating a reference vector along one direction $\mathbf{v}$ by angle $\theta$ according to the right hand rule \cite{rodrigues_lois_1840}. It is denoted as $\mathbf{d}=\mathbf{v}\tan{w/2}=(r_1,r_2,r_3)\in\mathcal{D}$, where $\|\mathbf{v}\|=1$ and $w\in[0,\pi]$.
\end{description}

Any of the aforementioned orientation representations have inherent ambiguity due to crystal symmetries. For example, if the crystal has cubic symmetry, its orientation is only uniquely defined up to a 48-fold set of rotations,  reflections and inversions of the cube about its symmetry axes. These rotations reflections, and inversions can be represented as a point symmetry group $\mathcal G$, called $m\overline{3}m$, of quaternionic matrices $\{\bQ_1, \ldots, \bQ_{48}\}$ operating on the 3D sphere.  Two orientations, e.g., represented in Euler angle, Quaternion or Rodrigues forms, are called symmetry-equivalent to each other if they are mapped to an equivalent orientation under $\mathcal G$. A fundamental zone (FZ), also called the fundamental domain, is a conic solid that can be specified to disambiguate any particular orientation $\bX_i$. However, as will be seen in Sec. \ref{sec:app_crystal_orientation_estimation}, reduction of the entire data sample $\{\bX_i\}_{i=1}^n$ to a FZ destroys information necessary for maximum likelihood estimation: the  entire $\mathcal G$-invariant density (\ref{eq:SSG}) must be used.

\subsection{$\mathcal G$-invariant Von-Mises Fisher distribution}

The von Mises-Fisher (VMF) distribution arises in directional statistics~\cite{mardia_directional_1999} as a natural generalization of the multivariate Gaussian distribution to the $(p-1)$-dimensional sphere $S^{(p-1)}\subset \Reals^p$, where $p\geq 2$.  The VMF distribution is parameterized by the mean direction $\bmu\in S^{(p-1)}$ and the concentration parameter $\kappa\ge 0$:
\begin{equation}
f(\mathbf{x};\mathbf{\bmu},\kappa) = c_p(\kappa)\exp{(\kappa\bmu^T\mathbf{x})},
\label{eq:VMF}
\end{equation}
where $c_p(\kappa) = \frac{\kappa^{p/2-1}}{(2\pi)^{p/2}I_{p/2-1}(\kappa)}$ and $I_p( \cdot)$ is the modified Bessel function of the first kind of order $p$. Given an i.i.d sample $\{\bX_i\}_{i=1}^n$ from the VMF distribution the ML estimator has the closed form expressions \cite{mardia_directional_1999}
\begin{align}
\label{eq:ML_estimator}
\hat{\bmu}=\frac{\mathbf{\gamma}}{\|\mathbf{\gamma}\|},\ \hat{\kappa}=A_p^{-1}\left(\frac{\|\mathbf{\gamma}\|}{n}\right),\
\end{align}
where $\mathbf{\gamma}=\sum_{i=1}^{n}\mathbf{\bx}_i$ and $A_p(u)=\frac{I_{p/2}(u)}{I_{p/2-1}(u)}$.

Let $\mathcal G$ be a group of symmetric actions $\{\bQ_1, \ldots, \bQ_M\}$ acting on the quaternionic representation of orientation on the $p-1$ dimensional sphere $S^{(p-1)}$. This group is called a spherical symmetry group. We extend the the VMF distribution (\ref{eq:VMF}) using the mixture representation Thm~\ref{thm:1}:
\begin{eqnarray}
\label{eq:mixture_densityp}
f(\mathbf{x};\bmu,\kappa)
&=&\sum_{m=1}^M\frac{1}{M}f(\bQ_m\mathbf{x};\bmu, \kappa)\\
\label{eq:mixture_density}
&=&\sum_{m=1}^M\frac{1}{M}f(\mathbf{x};\bQ_m\bmu, \kappa)
\end{eqnarray}
where in going from (\ref{eq:mixture_densityp}) to (\ref{eq:mixture_density}) we used the inner product form $\bmu^T \bx$ in (\ref{eq:VMF}) and the symmetry of $\bQ_m$. The expression (\ref{eq:mixture_density}) for the extended VMF distribution is in the form of
a finite mixture of standard VMF distributions on the same random variable $\bX$ having different mean parameters $\bmu_m =\bQ_m\bmu$ but having the same concentration parameters $\kappa$.

The finite mixture (\ref{eq:mixture_density}) for the $\mathcal G$-invariant density $f(\mathbf{x};\bmu,\kappa)$ is in a form for which an EM algorithm~\cite{dempster_maximum_1977} can be implemented to compute the ML estimates of $\bmu$ and $\kappa$.
Denoting the parameter pair as $\bomega=\{\bmu,\kappa\}$  the EM algorithm generates a sequence $\{\bomega_k\}_k$ of estimates that monotonically increase the likelihood and are given by $\bomega_{k+1}= \argmax_{\bomega}  E_{S|X,\bomega_k}[\log{L(\bomega;\{\bX_i,S_i\})}]$, where $S_i$ is a latent variable assigning $\bX_i$ to a particular mixture component in (\ref{eq:mixture_density}) and $L(\bomega,\{\bX_i, S_i\})$ is the likelihood function of $\bomega$ given the complete data $\{\bX_i, S_i\}_{i=1}^n$. Specifically,
\begin{eqnarray}
\label{eq:qfunction}
&&E_{S|X,\bomega}[\log{L(\bomega;\{\bX_i,S_i\})}] \\
&=& \sum_{i=1}^{n}\sum_{m=1}^Mr_{i,m}(\log{c_p(\kappa)}+\kappa(\bQ_m\bmu)^T\mathbf{x}_i),
\nonumber
\end{eqnarray}
where $r_{i,m}=P(S_i=m|\mathbf{X}_i;\bomega_k)$. The EM algorithm takes the form:

E-step:
\begin{equation}
\label{eq:EM_Estep}
\begin{split}
r_{i,m} 
&= \frac{f(\mathbf{X}_i; \bQ_m\bmu, \kappa)}{\sum_{l=1}^Mf(\mathbf{X}_i; \bQ_l\bmu, \kappa)}.
\end{split}
\end{equation}

M-step:
\begin{equation}
\label{eq:EM_Mstep}
\hat{\bmu}=\frac{\bgamma}{\|\bgamma\|},\ \hat{\kappa}=A_p^{-1}\left(\frac{\|\bgamma\|}{n}\right),
\end{equation}
where $\bgamma=\sum_{i=1}^{n}\sum_{m=1}^Mr_{i,m}\bQ_m^T\mathbf{X}_i$.

\section{Crystallographic Orientation Estimation}
\label{sec:app_crystal_orientation_estimation}
Polycrystalline materials are composed of grains, of varying size and orientation, where each grain contains crystal forms with similar orientations.  The quality of the material is mainly determined by the grain structure i.e. the arrangement of the grains and their sizes, as well as the distribution of the precipitates. Analyzing the crystal orientation of the grains helps us predict how materials fail and what modes of failure are more likely to occur \cite{de_graef_structure_2008}. 

Electron backscatter diffraction (EBSD) microscopy acquires crystal orientation at multiple locations within a grain by capturing the Kikuchi diffraction patterns of the backscatter electrons ~\cite{saruwatari_crystal_2007}. A Kikuchi pattern can be translated to crystal orientation through Hough Transformation analysis~\cite{lassen_automated_1994} or Dictionary-Based indexing~\cite{park_ebsd_2013}. The process of assigning mean orientation values to each grain is known as indexing. Crystal forms  possess  point symmetries, e.g. triclinic, tetragonal, or cubic, leading to  a probability density of measured orientations that is invariant over an associated spherical symmetry group $\mathcal G$. Therefore, when the type of material has known symmetries, e.g., cubic-type symmetry for nickel or gold, the extended VMF model introduced in the previous section can be applied to estimate the mean orientation $\bmu_g$ and the concentration $\kappa_g$ associated with each grain.

\subsection{Simulation studies of $\mathcal G$-invariant EM-ML estimator}
\begin{figure}[htb]
\centering
\centerline{\includegraphics[width=7cm]{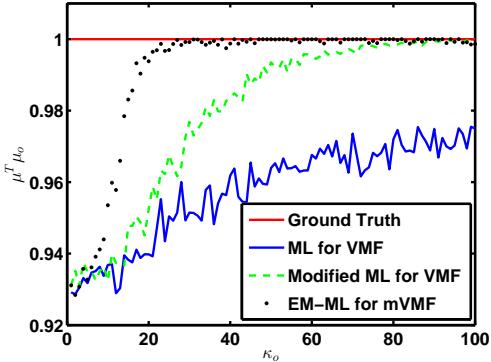}}
\caption{Mean orientation estimator comparisons for the $\mathcal G$-invariant density  when $\mathcal G$ is the $m \overline{3}m$ point symmetry group. Shown is the average inner product $\bmu_o^T\hat{\bmu}$ of three estimators $\hat{\bmu}$ when $\bmu_o$ is the true mean orientation as a function of the true concentration parameter $\kappa_o$. Each estimator was implemented with $n=1000$ i.i.d. samples from the $\mathcal G$-invariant density and the inner product shown is averaged over $100$ trials. The naive estimator ("ML for VMF" in blue line) does not attain perfect estimation (inner product $=1$) for any $\kappa_o$ since it does not account for the spherical symmetry group structure. A modified ML estimator  ("modified ML for VMF" in green dashed line) achieves perfect estimation as $\kappa_o$ becomes large.  The proposed EM-ML method ("EM-ML for mVMF") achieves perfect estimation much faster than the other methods.}
\label{fig:mu_estimation}
\end{figure}

\begin{figure}[htb]
\centering
\centerline{\includegraphics[width=7cm]{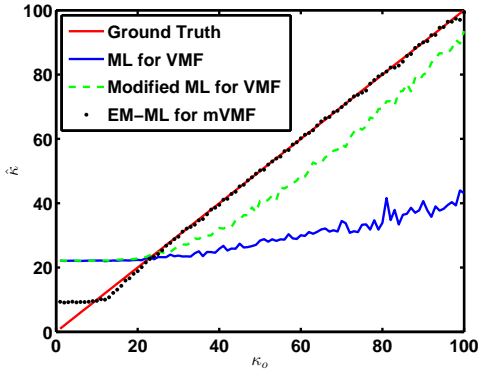}}
\caption{Concentration parameter estimator bias as a function of the  true concentration $\kappa_o$. The bias of the naive ML for VMF (blue solid line) is large over the full range of $\kappa_o$. The modified ML for VMF (green dashed line) estimates $\kappa$ more accurately when $\kappa_o$ is small. Our proposed method EM-ML estimator (black dotted line) has lower bias than the other estimators.}
\label{fig:kappa_estimation}
\end{figure}

A set of $n$ i.i.d. samples were simulated from the $\mathcal G$-invariant  VMF distribution with given $\bmu=\bmu_o,\kappa=\kappa_o$ for the $m\overline{3}m$ point symmetry group associated with the symmetries of cubic crystal lattice planes. The number of samples for each simulation was set to $n=1000$ and $\kappa_o$ was swept from  $1$ to $100$ while, for each simulation run,  $\bmu_o$ was selected uniformly at random. The experiment was repeated $100$ times and the average values of $\hat{\kappa}$ and the inner product $\hat{\bmu}^T \bmu_o$  are shown in Fig. \ref{fig:kappa_estimation} and \ref{fig:mu_estimation}. In the figures we compare performance for the following methods: (1) the naive ML estimator for the standard VMF model that does not account for the point group structure (\ref{eq:ML_estimator}) (labeled "ML for VMF"). (2) Mapping the $n$ samples to a single fundamental zone of $m\overline{3}m$ on the sphere followed by performing ML for the standard VMF distribution over this FZ (labeled "Modified ML for VMF"). (3) Applying our proposed exact EM-ML algorithm directly to the $n$ samples using the the mixture of VMF distribution (\ref{eq:EM_Estep})-(\ref{eq:EM_Mstep}) (labeled "EM-ML for mVMF").

Figure \ref{fig:mu_estimation} shows the inner product values $\bmu_o^T\hat{\bmu}$. The proposed EM-ML estimator achieves perfect recovery of the mean orientation ($\bmu_o^T\hat{\bmu}=1$) much faster than the other methods as the concentration parameter $\kappa_o$ increases (lower dispersion of the samples about the mean). Notice that when $\kappa_o< 20$, none of the methods can accurately estimate the mean orientation. The reason is that when $\kappa_o$ is small the samples become nearly uniformly distributed over the sphere. The threshold $\kappa_o$ value at which performance starts to degrade depends on the point symmetry group. In Fig. \ref{fig:kappa_estimation} it is seen that the bias of the proposed EM-ML $\kappa$ estimator is significantly lower than that of the other methods compared.  While the modified ML for VMF performs better than the naive ML estimator for VMF, its bias is significantly worse bias than the proposed EM-ML approach. 

\subsection{EM-ML orientation estimator for IN100 Nickel sample}
We next illustrate the proposed EM-ML orientation estimator on a real IN100 sample acquired from US Air Force Research Laboratory (AFRL) \cite{park_ebsd_2013}. The IN100 sample is a polycrystalline Ni superalloy which has cubic symmetry in the $m\overline{3}m$ point symmetry group.  EBSD orientation measurements were aquired on a $512\times 384$ pixel grid, corresponding to spatial resolution of $297.7$ nm. The Kikuchi diffraction patterns were recorded on a $80\times 60$ photosensitive detector for each of the pixels. 

\begin{figure}[htb]
\begin{minipage}[b]{.48\linewidth}
  \centering
  \centerline{\includegraphics[width=3.2cm]{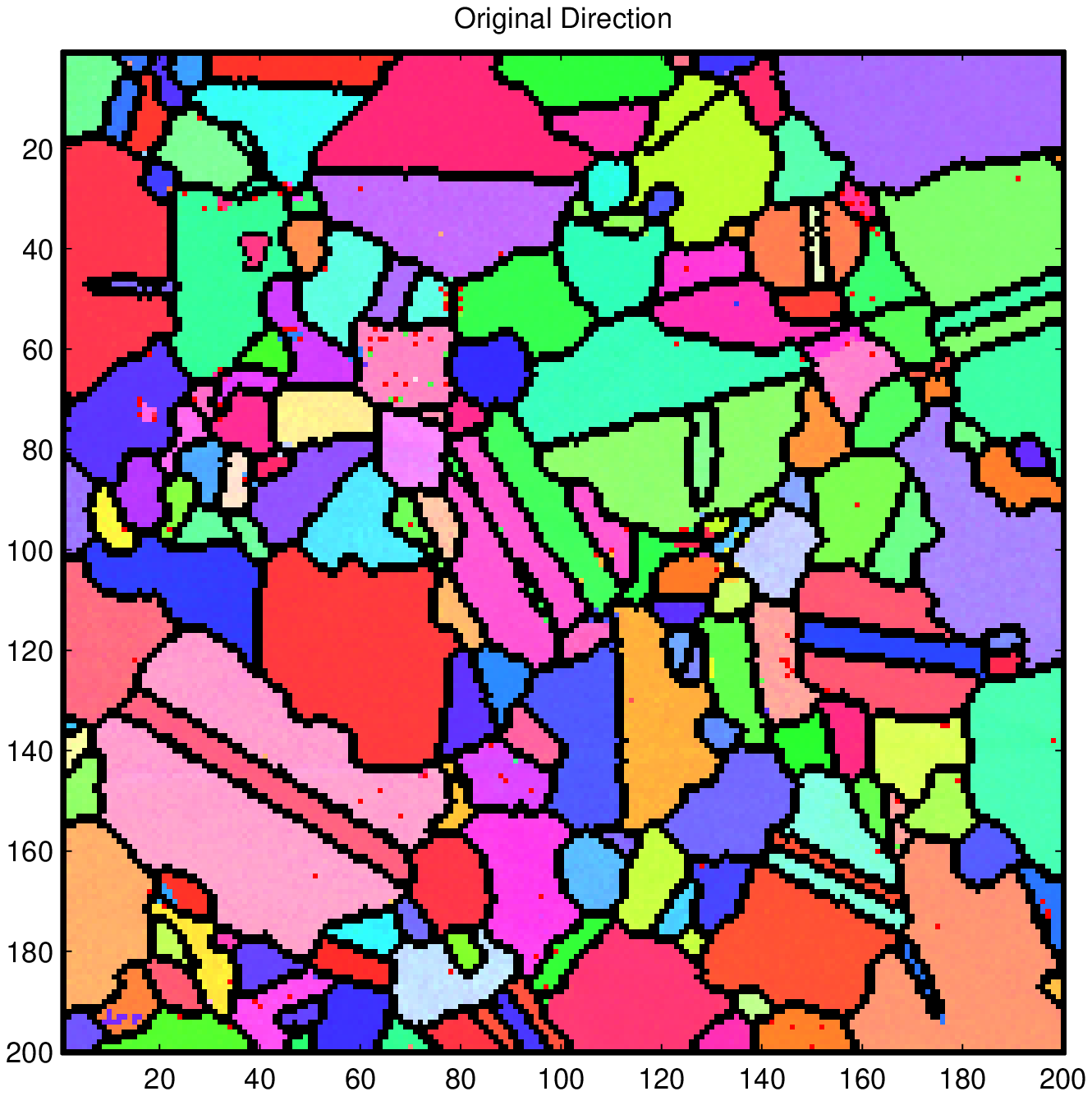}}
  \centerline{(a) IPF from OEM}\medskip
\end{minipage}
\begin{minipage}[b]{.48\linewidth}
  \centering
  \centerline{\includegraphics[width=3.2cm]{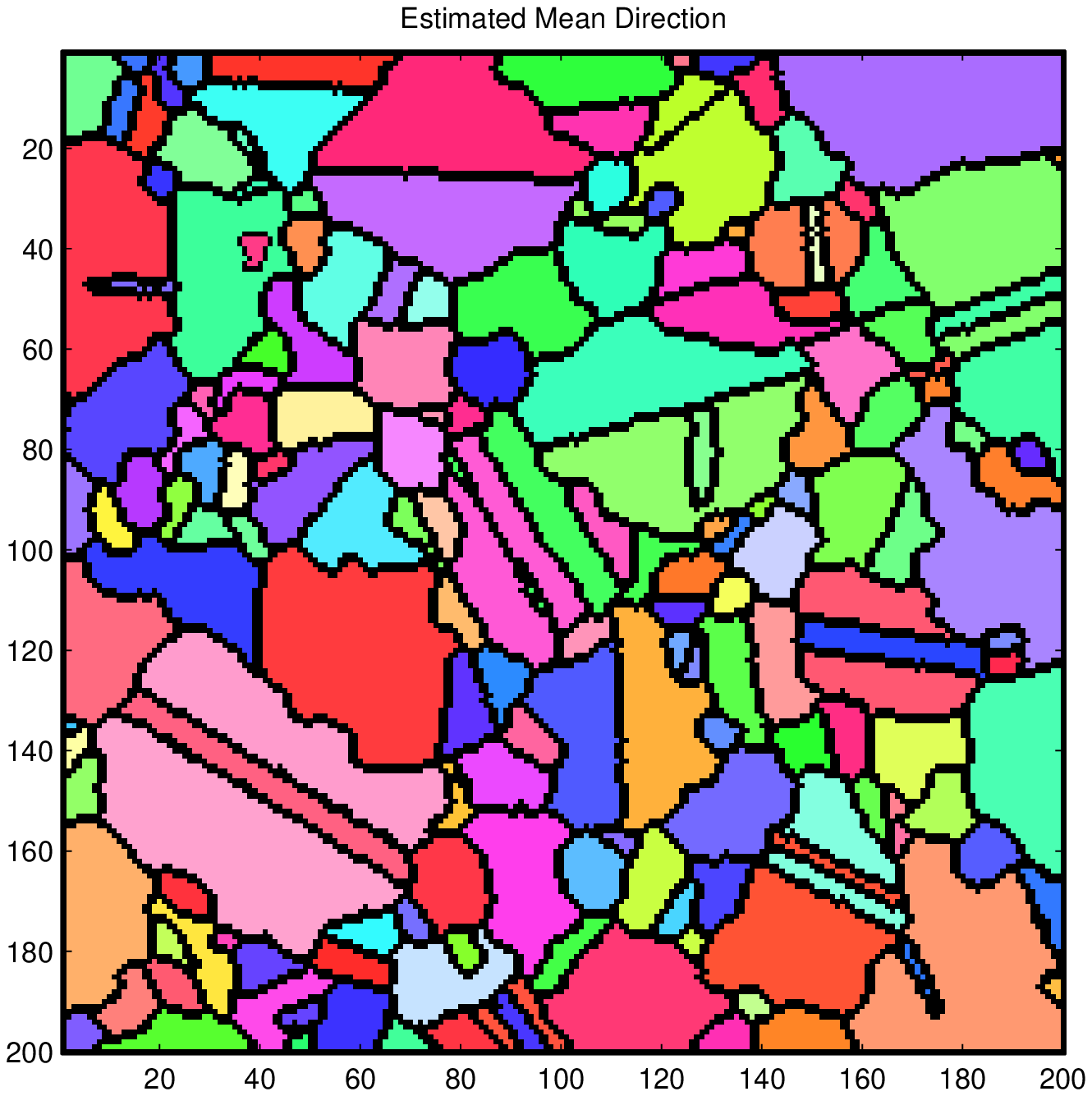}}
  \centerline{(b) IPF for proposed $\hat{\bmu}$}\medskip
\end{minipage}
\begin{minipage}[b]{.48\linewidth}
  \centering
  \centerline{\includegraphics[width=3.5cm]{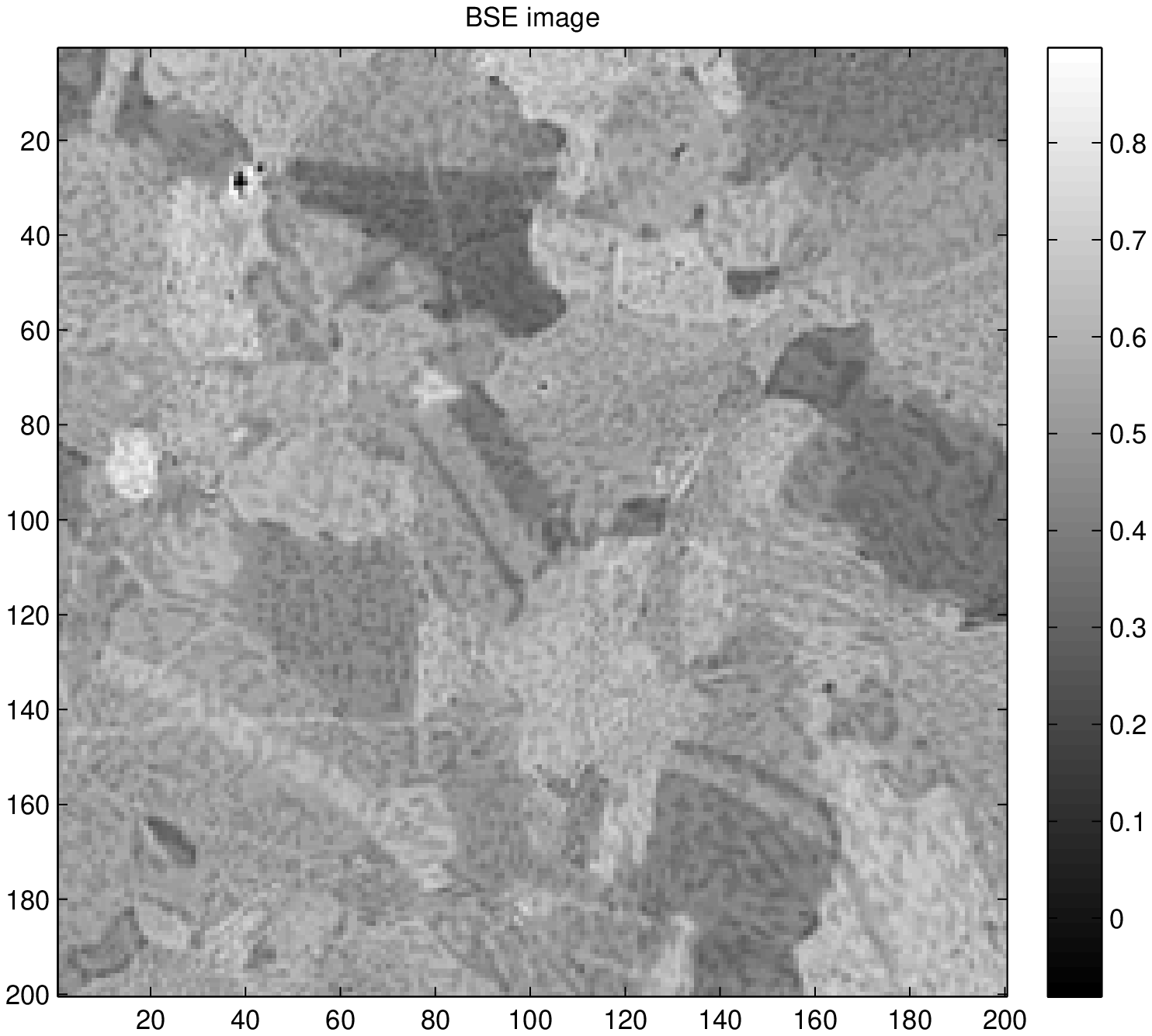}}
  \centerline{(c) BSE from OEM}\medskip
\end{minipage}
\hfill
\begin{minipage}[b]{0.48\linewidth}
  \centering
  \centerline{\includegraphics[width=3.5cm]{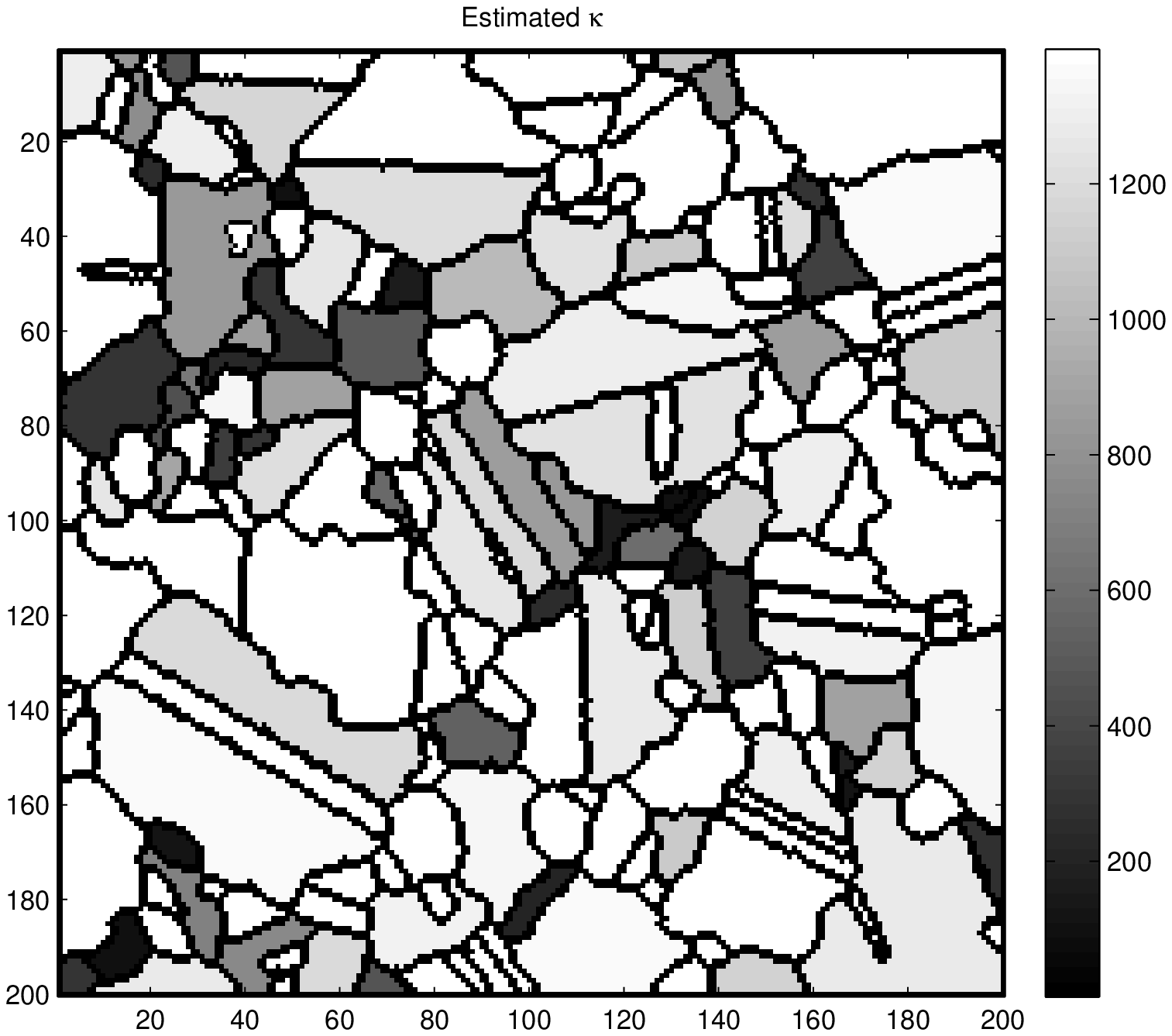}}
  \centerline{(d)$\hat{\kappa}$ for proposed $\hat{\bmu}$}\medskip
\end{minipage}
\caption{A $200\times 200$ sub-region of the IN100 sample. (a) is the IPF image for the Euler angles extracted from EBSD by OEM imaging software. IPF coloring in some grains is not homogeneous, likely due to the ambiguity problem. (b) is the IPF image for the mean orientation of the grains estimated by the proposed EM-ML algorithm. (c) is the BSE image of the sample and (d) is the concentration parameters $\kappa$ estimated by the proposed EM-ML for the $\mathcal G$-invariant VMF density. Our proposed EM-ML estimator has high concentration $\kappa$ even for those grains with inhomogeneous Euler angles than does the naive ML estimator.}
\label{fig:IN100}
\end{figure}

Figure \ref{fig:IN100} (a) shows a $200\times 200$ sub-region of the full EBSD sample where the orientations are shown in the inverse pole figure (IPF) coloring obtained from the OEM EBSD imaging software and (c) is the back-scattered electron (BSE) image. Note that the OEM-estimated orientations in some grain regions of the IPF image are very inhomogeneous, which is likely due to a fundamental zone wrap-around problem. Figure \ref{fig:IN100} (b) shows the estimates of the mean orientations of each grain using the proposed EM-ML algorithm. Figures \ref{fig:IN100} (d) show the estimated concentration parameter $\kappa$ for the grains using the proposed EM-ML algorithm. 

\section{Conclusion}
\label{sec:conclusion}
We have obtained a general finite mixture representation for densities on domains whose topologies have group invariances. This representation was used to extend the Von-Mises-Fisher distribution to a mixture VMF distributions that possess spherically symmetric group invariances. An efficient EM algorithm was derived for estimation of parameters of this extended VMF model. The extended VMF model was applied to the problem of estimation of mean grain orientation parameters in polycrystalline materials whose orientations lie in the $m\overline{3}m$ point symmetry group. Application of the finite mixture representation to other types of groups would be worthwhile future work.

\section*{Acknowledgment}
The authors are grateful for inputs from Megna Shah, Mike Jackson and Mike Groeber.

\bibliographystyle{IEEEtran}
\bibliography{SPL2015}

\newpage

 \ \newline
  
\newpage

\appendices
\section{Fundamental Zone for Cubic Symmetry}
\label{append:FZ_ROD}
In~\cite{morawiec_rodrigues_1996} a set of conditions defining the fundamental zone are given in terms of Rodrigues space:
\begin{equation}
\label{eq:union_FZ}
\bigcap_{i=2}^N\{r; tan(w_i/4)\pm r\mathbf{l}_i\ge0\}.
\end{equation}

Here $w_i\in[0,\pi]$ and $\mathbf{l}_i$ are the rotation angle and the unit vector of the rotation axis of the $i$-th element of the rotation symmetry group. $M$ is the order of the group and $i=1$ corresponds to the identity operator.

For cubic symmetry corresponding to the $m\overline{3}m$ point symmetry group, there are $24$ Rodrigues symmetry operators. The transformation equations between the Rodrigues vector $\mathbf{d}=(d_1, d_2, d_3)$ and the quaternion $\mathbf{q}=(q_1, q_2, q_3, q_4)$ are as follows:
\begin{equation}
\label{eq:transform_Rod_q}
q_1 = \frac{1}{\sqrt{1+\|\mathbf{d}\|^2}}; q_i=\frac{d_{i-1}}{\sqrt{1+\|\mathbf{d}\|^2}}, i\in[2,3,4].
\end{equation}

By applying Eq.(\ref{eq:union_FZ}) and Eq.(\ref{eq:transform_Rod_q}), the fundamental zone for the cubic structure in quaternion space obeys the following set of equations:
\begin{equation}
\label{eq:FZ_equations}
\begin{cases}
|q_2/q_1|\le\sqrt{2}-1 \\
|q_3/q_1|\le\sqrt{2}-1 \\
|q_4/q_1|\le\sqrt{2}-1 \\
|q_2/q_1 - q_3/q_1|\le\sqrt{2} \\
|q_2/q_1 + q_3/q_1|\le\sqrt{2} \\
|q_2/q_1 - q_4/q_1|\le\sqrt{2} \\
|q_2/q_1 + q_4/q_1|\le\sqrt{2} \\
|q_3/q_1 - q_4/q_1|\le\sqrt{2} \\
|q_3/q_1 + q_4/q_1|\le\sqrt{2} \\
|q_2/q_1 + q_3/q_1 + q_4/q_1|\le 1 \\
|q_2/q_1 - q_3/q_1 + q_4/q_1|\le 1 \\
|q_2/q_1 + q_3/q_1 - q_4/q_1|\le 1 \\
|q_2/q_1 - q_3/q_1 - q_4/q_1|\le 1
\end{cases}
\end{equation}

\section{EM Algorithm for Estimating von Mises-Fisher Distribution Parameters}
\label{append:EM_VMF}

The Expectation-Maximization algorithm is an iterative approach for obtaining maximum-likelihood parameter estimates in models where there are hidden latent variables and, in particular, finite mixture models. The algorithm alternates between performing two steps: "Expectation step (E-step)" and "Maximization step (M-step)" in each iteration. In the E-step, one calculates the expectation of the complete data log-likelihood function assuming the parameters of the model are fixed. In the M-step, the updated parameters are estimated by maximizing the expectation function. The process is repeated until the objective function converges. 

Here we assume that all the observed quaternions $X=\{\mathbf{x}_i\}_{i=1}^{n}$ are realizations from the $M$-fold finite mixture of Von Mises-Fisher (VMF) distributions which has the PDF (\ref{eq:mVMF_PDF_Append}).
\begin{equation}
\label{eq:mVMF_PDF_Append}
f_M(\mathbf{x};\bmu,\kappa)=\frac{1}{M}\sum_{m=1}^Mc_p(\kappa)\exp{(\kappa(\bQ_m\bmu)^T\mathbf{x})},
\end{equation}

Notice that The only parameters we need to estimate  are the mean parameter $\bmu$ and the concentration parameter $\kappa$. The latent variable $S_i\in[1, 2,..., M]$ indicates the index of the distribution a sample $\bX_i$, $i=1, \ldots, n$,  belongs to. Based on the model, the log-likelihood function given the data $\{\bX_i, S_i\}_{i=1}^n$  has the following form:
\begin{equation}
\label{eq:loglikelihood_EM_Append}
\begin{split}
& \log{L(\bomega;\{\bX_i,S_i\})} \nonumber \\
=& n\log{\frac{1}{M}}+\sum_{m=1}^M\sum_{i=1}^{n}(\log{c_p(\kappa)} + \kappa(\bQ_m\bmu)^T\mathbf{x}_i), \nonumber \\
\end{split}
\end{equation}
where $\bomega=\{\bmu,\kappa\}$ is the set of parameters. The EM objective function (called the $Q$ function) to be maximized is:
\begin{equation}
\label{eq:qfunction_Append}
\begin{split}
& Q(\bomega; \bomega_k) \\
=& E_{S|X,\bomega_k}[\log{L(\bomega;\{\bX_i,S_i\})}] \\
=& \sum_{i=1}^{n}\sum_{m=1}^Mr_{i,m}(\log{c_p(\kappa)}+\kappa(\bQ_m\bmu)^T\mathbf{x}_i),
\end{split}
\end{equation}
where $r_{i,m}$ is the posterior probability  that $S_i=m$. For the E-step, denote the parameters to be estimated as $\bomega_k=\{\bmu_k, \kappa_k\}$ at $k$-th iteration, $r_{i,m}$ can be calculated by:
\begin{equation}
\label{eq:EM_Estep_Append}
\begin{split}
r_{i,m} & = E[\mathds{1}(S_i=m)|\mathbf{x}_i; \bomega_k] \\
& = P(S_i=m | \mathbf{x}_i; \bomega_k) \\
& = \frac{c_p(\kappa_k)\exp{(\kappa_k(\bQ_m\bmu_k)^T\mathbf{x}_i)}}{\sum_{l=1}^Mc_p(\kappa_k)\exp{(\kappa_k(\bQ_l\bmu_k)^T\mathbf{x}_i)}}.
\end{split}
\end{equation}

In the M-step the parameters are updated by maximizing the $Q$ function. Taking the partial derivative w.r.t $\bmu$ of .(\ref{eq:qfunction_Append})) subject to the constraint $\|\bmu\|=1$ we have:
\begin{equation}
\begin{split}
&\frac{\partial}{\partial \bmu} Q(\bomega; \bomega_k) = \sum_{i=1}^{n}\sum_{m=1}^M r_{i,m}\kappa \bQ_m^T\mathbf{x}_i=2\lambda\bmu  \\
\Rightarrow&\hat{\bmu}=\frac{\bgamma}{\|\bgamma\|}, \bgamma=\sum_{i=1}^{n}\sum_{m=1}^Mr_{i,m}\bQ_m^T\mathbf{x}_i,
\end{split}
\end{equation}
where $\lambda$ is a Lagrange multiplier. By
taking the partial derivative w.r.t $\kappa$, the ML estimator of $\kappa$ is:
\begin{equation}
\begin{split}
&\frac{\partial}{\partial\kappa} Q(\bomega;\bomega_k) = N\frac{c_p^{'}(\kappa)}{c_p(\kappa)}+\kappa\bgamma^T\bmu \\
\Rightarrow&\hat{\kappa}=A_p^{-1}(\frac{\|\bgamma\|}{N}),
\end{split}
\end{equation}
where $A_p(u)=\frac{I_{p/2}(u)}{I_{p/2-1}(u)}$.

\end{document}